\documentclass[11pt,a4paper]{article}
\usepackage[hyperref]{acl2020}
\usepackage{times}
\usepackage{latexsym}

\usepackage[ruled]{algorithm2e}
\usepackage{soul}
\usepackage[utf8]{inputenc}
\usepackage{graphicx}
\usepackage{amsmath}
\usepackage{booktabs}
\usepackage{tabularx}
\usepackage{bbm}
\usepackage{amsfonts}
\usepackage{multirow}
\usepackage{subfig}
\usepackage{mathrsfs}
\usepackage{amsthm}
\usepackage{natbib}
\usepackage{amsmath,bm}
\usepackage{tablefootnote}
\usepackage{xcolor}

\usepackage{microtype}
\newtheorem{assumption}{Assumption}
\newtheorem{theorem}{Theorem}
\newcommand\blfootnote[1]{%
  \begingroup
  \renewcommand\thefootnote{}\footnote{#1}%
  \addtocounter{footnote}{-1}%
  \endgroup
}

\aclfinalcopy 
\def\aclpaperid{232} 

\title{Demographics Should Not Be the Reason of Toxicity: \\Mitigating Discrimination in Text Classifications with Instance Weighting}

\author{  
  Guanhua Zhang$^{\ddagger \S \ast}$, Bing Bai$^{\S \ast}$, Junqi Zhang$^\S$, Kun Bai$^\S$, Conghui Zhu$^\ddagger$, Tiejun Zhao$^\ddagger$\\
  $^\ddagger$Harbin Institute of Technology, $^\S$Tencent CSIG\\
  \texttt{\{guanhzhang,icebai,benjqzhang,kunbai\}@tencent.com},\\
  \texttt{\{chzhu,tjzhao\}@hit-mtlab.net}
}

\date{}

\makeatletter
\g@addto@macro{\UrlBreaks}{\UrlOrds}
\makeatother

\begin{document}
\maketitle

\begin{abstract}
  With the recent proliferation of the use of text classifications, researchers have found that there are certain unintended biases in text classification datasets.
  For example, texts containing some demographic identity-terms~(\emph{e.g.}, ``gay'', ``black'') are more likely to be abusive in existing abusive language detection datasets.
  As a result, models trained with these datasets may consider sentences like ``She makes me happy to be gay'' as abusive simply because of the word ``gay.''
  In this paper, we formalize the unintended biases in text classification datasets as a kind of \emph{selection bias} from the \emph{non-discrimination distribution} to the \emph{discrimination distribution}.
  Based on this formalization, we further propose a model-agnostic debiasing training framework by recovering the non-discrimination distribution using instance weighting, which does not require any extra resources or annotations apart from a pre-defined set of demographic identity-terms.
  Experiments demonstrate that our method can effectively alleviate the impacts of the unintended biases without significantly hurting models' generalization ability.
  \blfootnote{$^\ast$ Equal contributions from both authors. This work was done when Guanhua Zhang was an intern at Tencent.}
\end{abstract}

\section{Introduction}
\label{sec:intro}

With the development of Natural Language Processing~(NLP) techniques, Machine Learning~(ML) models are being applied in continuously expanding areas~(\emph{e.g.}, to detect spam emails, to filter resumes, to detect abusive comments), and they are affecting everybody's life from many aspects.
However, human-generated datasets may introduce some human social prejudices to the models~\cite{Caliskan2016Semantics}.
Recent works have found that ML models can capture, utilize, and even amplify the unintended biases~\cite{zhao2017men}, which has raised lots of concerns about the discrimination problem in NLP models~\cite{sun2019mitigating}.

Text classification is one of the fundamental tasks in NLP. It aims at assigning any given sentence to a specific class.
In this task, models are expected to make predictions with the semantic information rather than with the demographic group identity information~(\emph{e.g.}, ``gay'', ``black'') contained in the sentences.

However, recent research points out that there widely exist some unintended biases in text classification datasets.
For example, in a toxic comment identification dataset released by~\citet{dixon2018measuring}, it is found that texts containing some specific identity-terms are more likely to be toxic. More specifically, 57.4\% of comments containing ``gay'' are toxic, while only 9.6\% of all samples are toxic, as shown in Table~\ref{intr}.

\begin{table}[!t]
  \renewcommand\arraystretch{1.0}
  \centering
  \resizebox{0.43\textwidth}{!}{
    \begin{tabular}{c|p{2cm}<{\centering}|c}
      \hline
      Identity-term & Count   & Percentage Toxic \\
      \hline
      gay           & 868     & 57.4\%           \\
      homosexual    & 202     & 34.4\%           \\
      Mexican       & 116     & 21.6\%           \\
      blind         & 257     & 14.8\%           \\
      black         & 1,123   & 13.1\%           \\
      \hline
      overall       & 159,686 & 9.6\%            \\
      \hline
    \end{tabular}
  }
  \caption{Percentage of toxic comments by some specific demographic identity-terms in the dataset released by \citet{dixon2018measuring}.}
  \label{intr}
\end{table}


Because of such a phenomenon, models trained with the dataset may capture the unintended biases and perform differently for texts containing various identity-terms.
As a result, predictions of models may discriminate against some demographic minority groups. For instance, sentences like ``She makes me happy to be gay'' is judged as abusive by models trained on biased datasets in our experiment, which may hinder those minority groups who want to express their feelings on the web freely.

Recent model-agnostic research mitigating the unintended biases in text classifications can be summarized as data manipulation methods~\cite{sun2019mitigating}.
For example, \citet{dixon2018measuring} propose to apply data supplementation with additional labeled sentences to make toxic/non-toxic balanced across different demographic groups.
\citet{park2018reducing} proposes to use data augmentation by applying gender-swapping to sentences with identity-terms to mitigate gender bias.
The core of these works is to transform the training sets to an identity-balanced one. However, data manipulation is not always practical. Data supplementation often requires careful selection of the additional sentences \emph{w.r.t.} the identity-terms, the labels, and even the lengths of sentences~\cite{dixon2018measuring}, bringing a high cost for extra data collection and annotation.
Data augmentation may result in meaningless sentences~(\emph{e.g.}, ``He gives birth.''), and is impractical to perform when there are many demographic groups~(\emph{e.g.}, for racial bias cases).

In this paper, we propose a model-agnostic debiasing training framework that does not require any extra resources or annotations, apart from a pre-defined set of demographic identity-terms. We tackle this problem from another perspective, in which we treat the unintended bias as a kind of \emph{selection bias}~\cite{heckman1979sample}. We assume that there are two distributions, the \emph{non-discrimination distribution}, and the \emph{discrimination distribution} observed in the biased datasets, and every sample of the latter one is drawn independently from the former one following a discrimination rule, \emph{i.e.}, the social prejudice. With such a formalization, mitigating the unintended biases is equivalent to recovering the \emph{non-discrimination distribution} from the selection bias.
With a few reasonable assumptions, we prove that we can obtain the unbiased loss of the \emph{non-discrimination distribution} with only the samples from the observed \emph{discrimination distribution} with instance weights.
Based on this, we propose a non-discrimination learning framework.
Experiments on three datasets show that, despite requiring no extra data, our method is comparable to the data manipulation methods in terms of mitigating the discrimination of models.

The rest of the paper is organized as follows.
We summarize the related works in Section~\ref{sec:related_works}.
Then we give our perspective of the problem and examine the assumptions of commonly-used methods in Section~\ref{sec:perspective}.
Section~\ref{sec:method} introduces our non-discrimination learning framework.
Taking three datasets as examples, we report the experimental results of our methods in Section~\ref{sec:experiments}.
Finally, we conclude and present the future works in Section~\ref{sec:conclusion}.

\section{Related Works}
\label{sec:related_works}

\paragraph{Non-discrimination and Fairness}
Non-discrimination focuses on a number of protected demographic groups, and ask for parity of some statistical measures across these groups~\cite{chouldechova2017fair}.
As mentioned by \citet{friedler2016possibility}, non-discrimination can be achieved only if all groups have similar abilities \emph{w.r.t.} the task in the constructed space which contains the features that we would like to make a decision.
There are various kinds of definitions of non-discrimination corresponding to different statistical measures.
Popular measures include raw positive classification rate~\cite{calders2010three}, false positive and false negative rate~\cite{hardt2016equality} and positive predictive value~\cite{chouldechova2017fair}, corresponding to different definitions of non-discrimination.
Methods like adversarial training~\cite{beutel2017data,zhang2018mitigating} and fine-tuning~\cite{park2018reducing} have been applied to remove biasedness.

In the NLP area, fairness and discrimination problems have also gained tremendous attention.
\citet{Caliskan2016Semantics} show that semantics derived automatically from language corpora contain human biases.
\citet{bolukbasi2016man} show that pre-trained word embeddings trained on large-scale corpus can exhibit gender prejudices and provide a methodology for removing prejudices in embeddings by learning a gender subspace.
\citet{zhao2018gender} introduce the gender bias problem in coreference resolution and propose a general-purpose method for debiasing.

As for text classification tasks, \citet{dixon2018measuring} first points out the unintended bias in datasets and proposes to alleviate the bias by supplementing external labeled data.
\citet{kiritchenko2018examining} examines gender and race bias in 219 automatic sentiment analysis systems and finds that several models show significant bias.
\citet{park2018reducing} focus on the gender bias in abusive language detection task and propose to debias by augmenting the datasets with gender-swapping operation.
In this paper, we propose to make models fit a \emph{non-discrimination distribution} with calculated instance weights.

\paragraph{Instance Weighting}
Instance weighting has been broadly adopted for reducing bias.
For example, the Inverse~Propensity~Score~(IPS)~\cite{rosenbaum1983central} method has been successfully applied for causal effect analyses~\cite{austin2015moving}, selection bias~\cite{schonlau2009selection}, position bias~\cite{wang2018position,joachims2017unbiased} and so on.
\citet{zadrozny2004learning} proposed a methodology for learning and evaluating classifiers under ``Missing at Random''~(MAR)~\cite{rubin1976inference} selection bias. \citet{zhang2019selection} study the selection bias in natural language sentences matching datasets, and propose to fit a leakage-neutral distribution with instance weighting. \citet{jiang2007instance} propose an instance weighting framework for domain adaptation in NLP, which requires the data of the target domain.

In our work, we formalize the discrimination problem as a kind of ``Not Missing at Random''~(NMAR)~\cite{rubin1976inference} selection bias from the \emph{non-discrimination distribution} to the \emph{discrimination distribution}, and propose to mitigate the unintended bias with instance weighting.

\section{Perspective}
\label{sec:perspective}
In this section, we present our perspective regarding the discrimination problem in text classifications.
Firstly, we define what the \emph{non-discrimination distribution} is.
Then, we discuss what requirements non-discrimination models should meet and examine some commonly used criteria for non-discrimination. After that, we analyze some commonly used methods for assessing discrimination quantitatively.
Finally, we show that the existing debiasing methods can also be seen as trying to recover the \emph{non-discrimination distribution} and examine their assumptions.

\subsection{Non-discrimination Distribution}
\label{sec:definition_d_hat}

The unintended bias in the datasets is the legacy of the human society where discrimination widely exists. We denote the distribution in the biased datasets as \textbf{\emph{discrimination distribution} $\mathscr{D}$}.

Given the fact that the real world is discriminatory although it should not be, we assume that there is an ideal world where no discrimination exists, and the real world is merely a biased reflection of the non-discrimination world.
Under this perspective, we assume that there is an \textbf{\emph{non-discrimination distribution} $\widehat{\mathscr{D}}$} reflecting the ideal world, and the \emph{discrimination distribution} $\mathscr{D}$ is drawn from $\widehat{\mathscr{D}}$ but following a discriminatory rule, the social prejudice.
Attempting to correct the bias of datasets is equivalent to recover the original \emph{non-discrimination distribution} $\widehat{\mathscr{D}}$.

For the text classification tasks tackled in this paper, we denote $X$ as the sentences, $Y$ as the (binary) label indicator variable\footnote{In this paper, we focus on binary classification problems, but the proposed methodology can be easily extended to multi-class classifications.}, $Z$ as the demographic identity information~(e.g. ``gay'', ``black'', ``female'') in every sentence.
In the following paper, we use $P(\cdot)$ to represent the probability of the \emph{discrimination distribution} $\mathscr{D}$ in datasets, and $Q(\cdot)$ for \emph{non-discrimination distribution} $\widehat{\mathscr{D}}$.
Then the non-discrimination distribution $\widehat{\mathscr{D}}$ should meet that,
\begin{equation*}
  Q(Y|Z) = Q(Y)\,,
\end{equation*}
which means that the demographic identity information is independent of the labels\footnote{There may be a lot of distributions satisfying the equation. However, as we only focus on the discrimination problem in the text classification task, we suppose that there is a unique \emph{non-discrimination distribution} $\widehat{\mathscr{D}}$ which reflects the ideal world in the desired way and the observed biased dataset is drawn from it following a discriminatory rule.}.

\subsection{Non-Discrimination Model}
\label{sec:definition_eo}
For text classification tasks, models are expected to make predictions by understanding the semantics of sentences rather than by some single identity-terms.
As mentioned in \citet{dixon2018measuring}, a model is defined as biased if it performs better for sentences containing some specific identity-terms than for ones containing others.
In other words, a non-discrimination model should perform similarly across sentences containing different demographic groups. However, ``perform similarly'' is indeed hard to define. Thus, we pay more attention to some criteria defined on demographic groups.

A widely-used criterion is \emph{Equalized Odds}~(also known as \emph{Error Rate Balance}) defined by \citet{hardt2016equality}, requiring the $\widehat{Y}$ to be independent of $Z$ when $Y$ is given, in which $\widehat{Y}$ refers to the predictions of the model.
This criterion is also used by \citet{borkan2019nuanced} in text classifications.

Besides the \emph{Equalized Odds} criterion, a straightforward criterion for judging non-discrimination is \emph{Statistical Parity}~(also known as \emph{Demographic Parity}, \emph{Equal Acceptance Rates}, and \emph{Group Fairness})~\cite{calders2010three,dwork2012fairness}, which requires $\widehat{Y}$ to be independent of $Z$, \emph{i.e.}, $\text{Pr}(\widehat{Y}|Z)=\text{Pr}({\widehat{Y}})$.
Another criterion is \emph{Predictive Parity}~\cite{chouldechova2017fair}, which requires $Y$ to be independent of $Z$ when condition $\widehat{Y}=1$ is given, \emph{i.e.}, $\text{Pr}(Y|\widehat{Y}=1,Z)=\text{Pr}(Y|\widehat{Y}=1)$.
Given the definitions of the three criterions , we propose the following theorem, and the proof is presented in Appendix~\ref{app:proof_thm_cc}.

\begin{theorem}[Criterion Consistency]
  \label{thm:eq}
  When tested in a distribution in which $\text{Pr}(Y|Z)=\text{Pr}(Y)$, $\widehat{Y}$ satisfying Equalized Odds also satisfies Statistical Parity and Predictive Parity.
\end{theorem}

Based on the theorem, in this paper, we propose to evaluate models under a distribution where the demographic identity information is not predictive of labels to unify the three widely-used criteria.
Specifically, we define that a \emph{non-discrimination model} should meet that,
\begin{equation*}
  \text{Pr}(\widehat{Y}|Y, Z) = \text{Pr}(\widehat{Y}|Y)\,,
\end{equation*}
when tested in a distribution where $\text{Pr}(Y|Z)=\text{Pr}(Y)$.

\subsection{Assessing the Discrimination}
\label{sec:assessing}
Identity Phrase Templates Test Sets~(IPTTS) are widely used as non-discrimination testing sets to assess the models' discrimination~\cite{dixon2018measuring,park2018reducing,sun2019mitigating,kiritchenko2018examining}.
These testing sets are generated by several templates with slots for each of the identity-terms.
Identity-terms implying different demographic groups are slotted into the templates, \emph{e.g.}, ``I am a boy.'' and ``I am a girl.'', and it's easy to find that IPTTS satisfies $\text{Pr}(Y|Z)=\text{Pr}(Y)$.
A non-discrimination model is expected to perform similarly in sentences generated by the same template but with different identity-terms.

For metrics, False Positive Equality Difference~(FPED) and False Negative Equality Difference~(FNED) are used~\cite{dixon2018measuring,park2018reducing}, as defined below.
\begin{equation*}
  \begin{split}
    \text{FPED} &= \sum_{z}{|\text{FPR}_z-\text{FPR}_{overall}|}\,\text{,}\\
    \text{FNED} &= \sum_{z}{|\text{FNR}_z-\text{FNR}_{overall}|}\,\text{,}
  \end{split}
\end{equation*}
in which, $\text{FPR}_{overall}$ and $\text{FNR}_{overall}$, standing for False Positive Rate and False Negative Rate respectively, are calculated in the whole IPTTS.
Correspondingly, FPR$_z$ and FNR$_z$ are calculated on each subset of the data containing each specific identity-term.
These two metrics can be seen as a relaxation of \emph{Equalized Odds} mentioned in Section~\ref{sec:definition_eo}~\cite{borkan2019nuanced}.

It should also be emphasized that FPED and FNED do not evaluate the accuracy of models at all, and models can get lower FPED and FNED by making trivial predictions.
For example, when tested in a distribution where $\text{Pr}(Y|Z) = \text{Pr}(Y)$, if a model makes the same predictions for all inputs, FPED and FNED will be $0$, while the model is completely useless.

\subsection{Correcting the Discrimination}
\label{sec:correct}
Data manipulation has been applied to correct the discrimination in the datasets~\cite{sun2019mitigating}.
Previous works try to supplement or augment the datasets to an identity-balanced one, which, in our perspective, is primarily trying to recover the \emph{non-discrimination distribution} $\widehat{\mathscr{D}}$.

For data supplementation, \citet{dixon2018measuring} adds some additional non-toxic samples containing those identity-terms which appear disproportionately across labels in the original biased dataset.
Although the method is reasonable, due to high cost, it is not always practical to add additional labeled data with specific identity-terms, as careful selection of the additional sentences \emph{w.r.t.} the identity-terms, the labels, and even the lengths of sentences is required~\cite{dixon2018measuring}.

The gender-swapping augmentation is a more common operation to mitigate the unintended bias~\cite{zhao2018gender,sun2019mitigating}.
For text classification tasks, \citet{park2018reducing} augment the datasets by swapping the gender-implying identity-terms~(\emph{e.g.}, ``he'' to ``she'', ``actor'' to ``actress'') in the sentences of the training data to remove the correlation between $Z$ and $Y$.
However, it is worth mentioning that the gender-swapping operation additionally assumes that the \emph{non-discrimination distribution} $\widehat{\mathscr{D}}$ meets the followings,
\begin{equation*}
  \begin{split}
    Q(X^\lnot|Z) &= Q(X^\lnot)\,\text{,}\\
    Q(Y|X^\lnot,Z) &= Q(Y|X^\lnot)\,\text{,}
  \end{split}
\end{equation*}
in which $X^\lnot$ refers to the content of sentences except for the identity information.
And we argue that these assumptions may not hold sometimes.
For example, the first assumption may result in some meaningless sentences~(\emph{e.g.}, \emph{``He gives birth.''})~\cite{sun2019mitigating}.
Besides, this method is not practical for situations with many demographic groups.

\section{Our Instance Weighting Method}
\label{sec:method}

In this section, we introduce the proposed method for mitigating discrimination in text classifications. We first make a few assumptions about how the \emph{discrimination distribution} $\mathscr{D}$ in the datasets are generated from the \emph{non-discrimination distribution} $\widehat{\mathscr{D}}$. Then we demonstrate that we can obtain the unbiased loss on $\widehat{\mathscr{D}}$ only with the samples from $\mathscr{D}$, which makes models able to fit the \emph{non-discrimination distribution}  $\widehat{\mathscr{D}}$ without extra resources or annotations.


\subsection{Assumptions about the Generation Process}
Considering the perspective that the \emph{discrimination distribution} is generated from the \emph{non-discrimination distribution} $\widehat{\mathscr{D}}$, we refer $S\in[0, 1]$ as the selection indicator variable, which indicates whether a sample is selected into the biased dataset or not.
Specifically, we assume that every sample $(x, z, y, s)$\footnote{Definitions of $x$, $z$ and $y$ are in Section~\ref{sec:definition_d_hat}.} is drawn independently from $\widehat{\mathscr{D}}$ following the rule that, if $s=1$ then the sample is selected into the dataset, otherwise it is discarded, then we have
\begin{assumption}
  $P(\cdot) = Q(\cdot|S=1)\,,$
\label{asmp:1}
\end{assumption}
\noindent and as defined in Section~\ref{sec:definition_d_hat}, the \emph{non-discrimination distribution} $\widehat{\mathscr{D}}$ satisfies
\begin{assumption}
  $Q(Y|Z) = Q(Y)\,.$
\label{asmp:2}
\end{assumption}

Ideally, if the values of $S$ are entirely at random, then the generated dataset can correctly reflect the original \emph{non-discrimination distribution} $\widehat{\mathscr{D}}$ and does not have discrimination.
However, due to social prejudices, the value of $S$ is not random.
Inspired by the fact that some identity-terms are more associated with some specific labels than other identity-terms~(\emph{e.g.}, sentences containing ``gay'' are more likely to be abusive in the dataset as mentioned before), we assume that $S$ is controlled by $Y$ and $Z$\footnote{
  As we only focus on the discrimination problem in this work, we ignore selection bias on other variables like \emph{topic} and \emph{domain}.
}.
We also assume that, given any $Z$ and $Y$, the conditional probability of $S=1$ is greater than $0$, defined as,
\begin{assumption}
  $Q(S=1|X, Y, Z) = Q(S=1|Y, Z) > 0\,.$
\label{asmp:3}
\end{assumption}

Meanwhile, we assume that the social prejudices will not change the marginal probability distribution of $Z$, defined as,
\begin{assumption}
  $P(Z) = Q(Z)\,,$
\label{asmp:4}
\end{assumption}
\noindent which also means that $S$ is independent with $Z$ in $\widehat{\mathscr{D}}$, \emph{i.e.}, $Q(S|Z)=Q(S)$. 

Among them, Assumption~\ref{asmp:1} and \ref{asmp:2} come from our problem framing. Assumption~\ref{asmp:3} helps simplify the problem. Assumption~\ref{asmp:4} helps establish the non-discrimination distribution $\widehat{\mathscr{D}}$. 
Theoretically, when $Z$ is contained in $X$, which is a common case, consistent learners should be asymptotically immune to this assumption~\cite{fan2005improved}.
A more thorough discussion about Assumption~\ref{asmp:4} can be found in Appendix~\ref{app:pzqz}.

\subsection{Making Models Fit the \emph{Non-discrimination Distribution} $\widehat{\mathscr{D}}$}

\paragraph{Unbiased Expectation of Loss}
Based on the assumptions above, we prove that we can obtain the loss unbiased to the \emph{non-discrimination distribution} $\widehat{\mathscr{D}}$ from the \emph{discrimination distribution} with calculated instance weights.

\begin{theorem}[Unbiased Loss Expectation]
  \label{thm}
  For any classifier $f=f(x, z)$, and for any loss function $\Delta = \Delta(f(x,z), y)$, if we use $w=\frac{Q(y)}{P(y|z)}$ as the instance weights, then

  \begin{small}
    \begin{displaymath}
      E_{x,y,z \sim \mathscr{D}} \Big[w\Delta\big(f(x,z), y\big) \Big] = E_{x, y, z \sim \widehat{\mathscr{D}}} \Big[ \Delta(f(x,z), y) \Big] \,.
    \end{displaymath}
  \end{small}

\end{theorem}

Then we present the proof for Theorem~\ref{thm}.
\begin{proof}
  We first present an equation with the weight $w$, in which we use numbers to denote assumptions used in each step and \emph{bayes} for \emph{Bayes' Theorem}.
  \begin{scriptsize}
    \begin{equation*}
      \begin{split}
        w=&  \frac{Q(y)}{P(y|z)} \\
        \overset{1}{=}&  \frac{Q(y)}{Q(y|z, S=1)} \\
        \overset{bayes}{=}&\frac{Q(y)}{Q(S=1|z,y) Q(y|z) / Q(S=1|z)}\\
        \overset{2,4}{=}&  \frac{Q(S=1)}{Q(S=1|z,y)}\\
        \overset{3, bayes}{=}&  \frac{Q(S=1)}{Q(x,z,y|S=1)Q(S=1) / Q(x,z,y)}\\
        \overset{1}{=}&  \frac{Q(x,z,y)}{P(x,z,y)}\,\text{.}
      \end{split}
    \end{equation*}
  \end{scriptsize}

  Then we have

  \begin{footnotesize}
    \begin{displaymath}
      \begin{split}
        & E_{x,z,y \sim \mathscr{D}} \Big[w\Delta\big(f(x,z), y\big) \Big] \\
        = & \int \frac{Q(x, z, y)}{P(x, z, y)} \Delta(f(x,z), y) dP(x,z,y) \\
        = & \int \Delta(f(x,z), y) dQ(x,z,y) \\
        = & E_{x, y, z \sim \widehat{\mathscr{D}}} \Big[ \Delta(f(x,z), y) \Big] \,.
      \end{split}
    \end{displaymath}
  \end{footnotesize}
\end{proof}

\paragraph{Non-discrimination Learning}
Theorem~\ref{thm} shows that, we can obtain the unbiased loss of the \emph{non-discrimination distribution} $\widehat{\mathscr{D}}$ by adding proper instance weights to the samples from the \emph{discrimination distribution} $\mathscr{D}$.
In other words, non-discrimination models can be trained with the instance weights $w=\frac{Q(y)}{P(y|z)}$.
As the \emph{discrimination distribution} is directly observable, estimating $P(y|z)$ is not hard.
In practice, we can train classifiers and use cross predictions to estimate $P(y|z)$ in the original datasets.
Since $Q(y)$ is only a real number indicating the prior probability of $Y \in [0, 1]$ on distribution $\widehat{\mathscr{D}}$, we do not specifically make an assumption on it. 
Intuitively, setting $Q(Y) = P(Y)$ can be a good choice.
Considering an non-discrimination dataset where $P(Y|Z)=P(Y)$, the calculated weights $\frac{Q(y)}{P(y|z)}$ should be the same for all samples when we set $Q(Y)=P(Y)$, and thus have little impacts on trained models.

We present the step-by-step procedure for non-discrimination learning in Algorithm 1. Note that the required data is only the biased dataset, and a pre-defined set of demographic identity-terms, with which we can extract $\{x, y, z\}$ for all the samples.

\begin{table}[t!]
  \renewcommand\arraystretch{1.0}
  \begin{center}
    \resizebox{0.482\textwidth}{!}{
      \begin{tabular}{p{0.15cm}<{\centering} p{8.5cm}}
        \hline
        \multicolumn{2}{l}{\textbf{Algorithm 1: Non-discrimination Learning}}                                                                                             \\
        \hline
        \multicolumn{2}{p{9.0cm}}{\textbf{Input}: The dataset $\{x, z, y\}$, the number of fold $K$ for cross prediction and the prior probability $Q(Y=0)$ and $Q(Y=1)$} \\
        \multicolumn{2}{l}{\textbf{Procedure}:}                                                                                                                           \\
        01 & Train classifiers and use $K$-fold cross-predictions to estimating $P(y|z)$ with the dataset                                                                 \\
        02 & Calculate the weights $w=\frac{Q(y)}{P(y|z)}$ for all samples                                                                                                \\
        03 & Train and validate models using $w$ as the instance weights                                                                                                  \\
        \hline
      \end{tabular}}
  \end{center}
\end{table}

\section{Experiments}
\label{sec:experiments}

In this section, we present the experimental results for non-discrimination learning.
We demonstrate that our method can effectively mitigate the impacts of unintended discriminatory biases in datasets.

\subsection{Dataset Usage}

We evaluate our methods on three datasets, including the Sexist Tweets dataset, the Toxicity Comments dataset, and the Jigsaw Toxicity dataset.

\paragraph{Sexist Tweets}
We use the Sexist Tweets dataset released by~\citet{waseem2016hateful,waseem2016you}, which is for abusive language detection task\footnote{Unfortunately, due to the rules of Twitter, some TweetIDs got expired, so we cannot collect the exact same dataset as~\citet{park2018reducing}.}.
The dataset consists of tweets annotated by experts as ``sexist'' or ``normal.''
We process the dataset as to how \citet{park2018reducing} does.
It is reported that the dataset has an unintended gender bias so that models trained in this dataset may consider ``You are a good woman.'' as ``sexist.''
We randomly split the dataset in a ratio of $8:1:1$ for training-validation-testing and use this dataset to evaluate our method's effectiveness on mitigating gender discrimination.

\paragraph{Toxicity Comments}
Another choice is the Toxicity Comments dataset released by~\citet{dixon2018measuring}, in which texts are extracted from Wikipedia Talk Pages and labeled by human raters as either toxic or non-toxic.
It is found that in this dataset, some demographic identity-terms~(\emph{e.g.}, ``gay'', ``black'') appear disproportionately among labels.
As a result, models trained in this dataset can be discriminatory among groups.
We adopt the split released by \citet{dixon2018measuring} and use this dataset to evaluate our method's effectiveness on mitigating discrimination towards minority groups.

\paragraph{Jigsaw Toxicity}
We also tested a recently released large-scale dataset Jigsaw Toxicity from Kaggle\footnote{https://www.kaggle.com/c/jigsaw-unintended-bias-in-toxicity-classification}, in which it is found that some frequently attacked identities are associated with toxicity.
Sentences in the dataset are extracted from the Civil Comment platform and annotated with toxicity and identities mentioned in every sentence.
We randomly split the dataset into $80\%$ for training, $10\%$ for validation and testing respectively.
The dataset is used to evaluate our method's effectiveness on large-scale datasets.

The statistic of the three datasets is shown as in Table~\ref{stat}.
\begin{table}[!t]
  \renewcommand\arraystretch{1.0}
  \centering
  \resizebox{0.475\textwidth}{!}{
    \begin{tabular}{c|c|c|c}
      \hline
      Dataset         & Size      & Positives & avg. Length \\
      \hline
      Sexist Tweets   & 12,097    & 24.7\%    & 14.7        \\
      Toxicity Comments  & 159,686   & 9.6\%     & 68.2        \\
      Jigsaw Toxicity & 1,804,874 & 8.0\%     & 51.3        \\
      \hline
    \end{tabular}
  }
  \caption{Statistics of the three datasets for evaluation.}
  \label{stat}
\end{table}

\subsection{Evaluation Scheme}
Apart from the original testing set of each dataset, we use the Identity Phrase Templates Test Sets~(IPTTS) described in Section~\ref{sec:assessing} to evaluate the models as mentioned in Section~\ref{sec:assessing}.
For experiments with the Sexist Tweets dataset, we generate IPTTS following~\citet{park2018reducing}.
For experiments with Toxicity Comments datasets and Jigsaw Toxicity, we use the IPTTS released by~\citet{dixon2018measuring}.
Details about the IPTTS generation are introduced in Apendix~\ref{app:iptts}.

For metrics, we use FPED and FNED in IPTTS to evaluate how discriminatory the models are, and lower scores indicate better equality.
However, as mentioned in Section~\ref{sec:assessing}, these two metrics are not enough since models can achieve low FPED and FNED by making trivial predictions in IPTTS.
So we use AUC in both the original testing set and IPTTS to reflect the trade-off between the debiasing effect and the accuracy of models.
We also report the significance test results under confidence levels of 0.05 for Sexist Tweets dataset and Jigsaw Toxicity dataset\footnote{As we use some results from \citet{dixon2018measuring} directly, we don't report the significance test results for Toxicity Comments dataset.}.

For baselines, we compare with the gender-swapping method proposed by~\citet{park2018reducing} for the Sexist Tweets dataset, as there are only two demographics groups (male and female) provided by the dataset,  it's practical for swapping. For the other two datasets, there are 50 demographics groups, and we compare them with data supplementation proposed by~\citet{dixon2018measuring}.

\subsection{Experiment Setup}
To generate the weights, we use Random Forest Classifiers to estimate $P(y|z)$ following Algorithm~1.
We simply set $Q(Y) = P(Y)$ to partial out the influence of the prior probability of $Y$.
The weights are used as the sample weights to the loss functions during training and validation.

For experiments with the Sexist Tweets dataset, we extract the gender identity words~(released by \citet{zhao2018gender}) in every sentence and used them as $Z$.
For experiments with Toxicity Comments dataset, we take the demographic group identity words~(released by \citet{dixon2018measuring}) contained in every sentence concatenated with the lengths of sentences as $Z$, just the same as how \citet{dixon2018measuring} chose the additional sentence for data supplement.
For experiments with the Jigsaw Toxicity dataset, the provided identity attributes of every sentence and lengths of sentences are used as $Z$.

For experiments with the Toxicity Comments dataset, to compare with the results released by \citet{dixon2018measuring}, we use their released codes, where a three-layer Convolutional Neural Network~(CNN) model is used.
For experiments with Sexist Tweets dataset and Jigsaw Toxicity dataset, as our method is model-agnostic, we simply implement a one-layer LSTM with a dimensionality of 128 using Keras and Tensorflow backend.\footnote{Codes are publicly available at \url{https://github.com/ghzhang233/Non-Discrimination-Learning-for-Text-Classification}.}

For all models, pre-trained GloVe word embeddings~\cite{pennington2014glove} are used.
We also report results when using gender-debiased pre-trained embeddings~\cite{bolukbasi2016man} for experiments with Sexist Tweets. All the reported results are the average numbers of ten runs with different random initializations.

\subsection{Experimental Results}
In this section, we present and discuss the experimental results.
As expected, training with calculated weights can effectively mitigate the impacts of the unintended bias in the datasets.

\begin{table}[!t]
  \renewcommand\arraystretch{1.0}
  \centering
  \resizebox{0.482\textwidth}{!}{
    \begin{tabular}{c|c|c|c|c}
      \hline
      Model        & Orig. AUC      & IPTTS AUC      & FPED           & FNED           \\
      \hline
      Baseline     & \textbf{0.920} & 0.673          & 0.147          & 0.204          \\
      Swap         & 0.911$\dagger$          & 0.651          & \textbf{0.047}$\dagger$ & \textbf{0.050}$\dagger$ \\
      Weight       & 0.897$\dagger\,\ddagger$          & \textbf{0.686}$\ddagger$ & 0.057$\dagger$          & 0.086$\dagger\,\ddagger$          \\
      \hline
      Baseline$^+$ & \textbf{0.900} & 0.624          & 0.049          & 0.099          \\
      Swap$^+$     & 0.890$\dagger$          & 0.611          & 0.008$\dagger$          & \textbf{0.013}$\dagger$ \\
      Weight$^+$   & 0.881$\dagger\,\ddagger$          & \textbf{0.647}$\dagger\,\ddagger$ & \textbf{0.007}$\dagger$ & 0.024$\dagger\,\ddagger$          \\
      \hline
      \multicolumn{5}{r}{\small $\dagger \ p<0.05$ compared with \emph{Baseline}}\\
      \multicolumn{5}{r}{\small $\ddagger \ p<0.05$ compared with \emph{Swap}}\\
    \end{tabular}
  }
  \vspace{-8pt}
  \caption{Experimental results with Sexist Tweets dataset. ``$+$'' refers to models using debiased word embeddings.}
  \label{tab5}
\end{table}

\begin{table}[!t]
  \renewcommand\arraystretch{1.0}
  \centering
  \resizebox{0.482\textwidth}{!}{
    \begin{tabular}{c|c|c|c|c}
      \hline
      Model      & Orig. AUC      & IPTTS AUC      & FPED           & FNED           \\
      \hline
      Baseline   & \textbf{0.960} & 0.952          & 7.413          & 3.673          \\
      Supplement & 0.959          & 0.960          & 5.294          & 3.073          \\
      Weight     & 0.956          & \textbf{0.961} & \textbf{4.798} & \textbf{2.491} \\
      \hline
       \multicolumn{5}{r}{\small The results of \emph{Baseline} and \emph{Supplement} are taken from \citet{dixon2018measuring}}\\
    \end{tabular}}
  \vspace{-5pt}
  \caption{Experimental results with Toxicity Comments dataset. }
  \label{tab4}
\end{table}

\begin{table}[!t]
  \renewcommand\arraystretch{1.0}
  \centering
  \resizebox{0.482\textwidth}{!}{
    \begin{tabular}{c|c|c|c|c}
      \hline
      Model      & Orig. AUC      & IPTTS AUC      & FPED           & FNED           \\
      \hline
      Baseline   & \textbf{0.928} & 0.993          & 3.088          & 3.317          \\
      Supplement & \textbf{0.928} & \textbf{0.999}$\dagger$ & 0.180$\dagger$          & 3.111          \\
      Weight     & 0.922$\dagger\,\ddagger$          & \textbf{0.999}$\dagger$ & \textbf{0.085}$\dagger$ & \textbf{2.538} \\
      \hline
      \multicolumn{5}{r}{\small $\dagger \ p<0.05$ compared with \emph{Baseline}}\\
      \multicolumn{5}{r}{\small $\ddagger \ p<0.05$ compared with \emph{Supplement}}\\
    \end{tabular}
  }
  \vspace{-8pt}
  \caption{Experimental results with Jigsaw Toxicity dataset.}
  \label{tab6}
\end{table}

\paragraph{Sexist Tweets}
Tabel~\ref{tab5} reports the results on Sexist Tweets dataset. \textbf{\emph{Baseline}} refers to vanilla models. \textbf{\emph{Swap}} refers to models trained and validated with 2723 additional gender-swapped samples to balance the identity-terms across labels~\cite{park2018reducing}. \textbf{\emph{Weight}} refers to models trained and validated with calculated weights. ``$+$'' refers to models using debiased word embeddings.

Regarding the results with the GloVe word embeddings, we can find that \emph{Weight} performs significantly better than \emph{Baseline} under FPED and FNED, which demonstrate that our method can effectively mitigate the discrimination of models.
\emph{Swap} outperforms \emph{Weight} in FPED and FNED, but our method achieves significantly higher IPTTS AUC.
We notice that \emph{Swap} even performs worse in terms of IPTTS AUC than \emph{Baseline}~(although the difference is not significant at 0.05), which implies that cost for the debiasing effect of \emph{Swap} is the loss of models' accuracy, and this can be ascribed to the gender-swapping assumptions as mentioned in Section~\ref{sec:correct}.
We also notice that both \emph{Weight} and \emph{Swap} have lower Orig. AUC than \emph{Baseline} and this can be ascribed to that the unintended bias pattern is mitigated.

Regarding the results with the debiased word embeddings, the conclusions remain largely unchanged, while \emph{Weight} get a significant improvement over \emph{Baseline} in terms of IPTTS AUC. Besides, compared with GloVe embeddings, we can find that debiased embeddings can effectively improve FPED and FNED, but Orig. AUC and IPTTS AUC also drop.

\paragraph{Toxicity Comments}
Table~\ref{tab4} reports the results on Toxicity Comments dataset. \textbf{\emph{Baseline}} refers to vanilla models. \textbf{\emph{Supplement}} refers to models trained and validated with $4620$ additional samples to balance the identity-terms across labels~\cite{dixon2018measuring}.
\textbf{\emph{Weight}} refers to models trained and validated with calculated instance weights.

From the table, we can find that \emph{Weight} outperforms \emph{Baseline}
in terms of IPTTS AUC, FPED, and FNED, and also gives sightly better debiasing performance compared with \emph{Supplement}, which demonstrate that the calculated weights can effectively make models more non-discriminatory.
Meanwhile, \emph{Weight} performs similarly in Orig. AUC to all the other methods, indicating that our method does not hurt models' generalization ability very much.

In general, the results demonstrate that our method can provide a better debiasing effect without additional data, and avoiding the high cost of extra data collection and annotation makes it more practical for adoptions.

\begin{figure}[!t]
  \centering
  \includegraphics[width=0.482\textwidth]{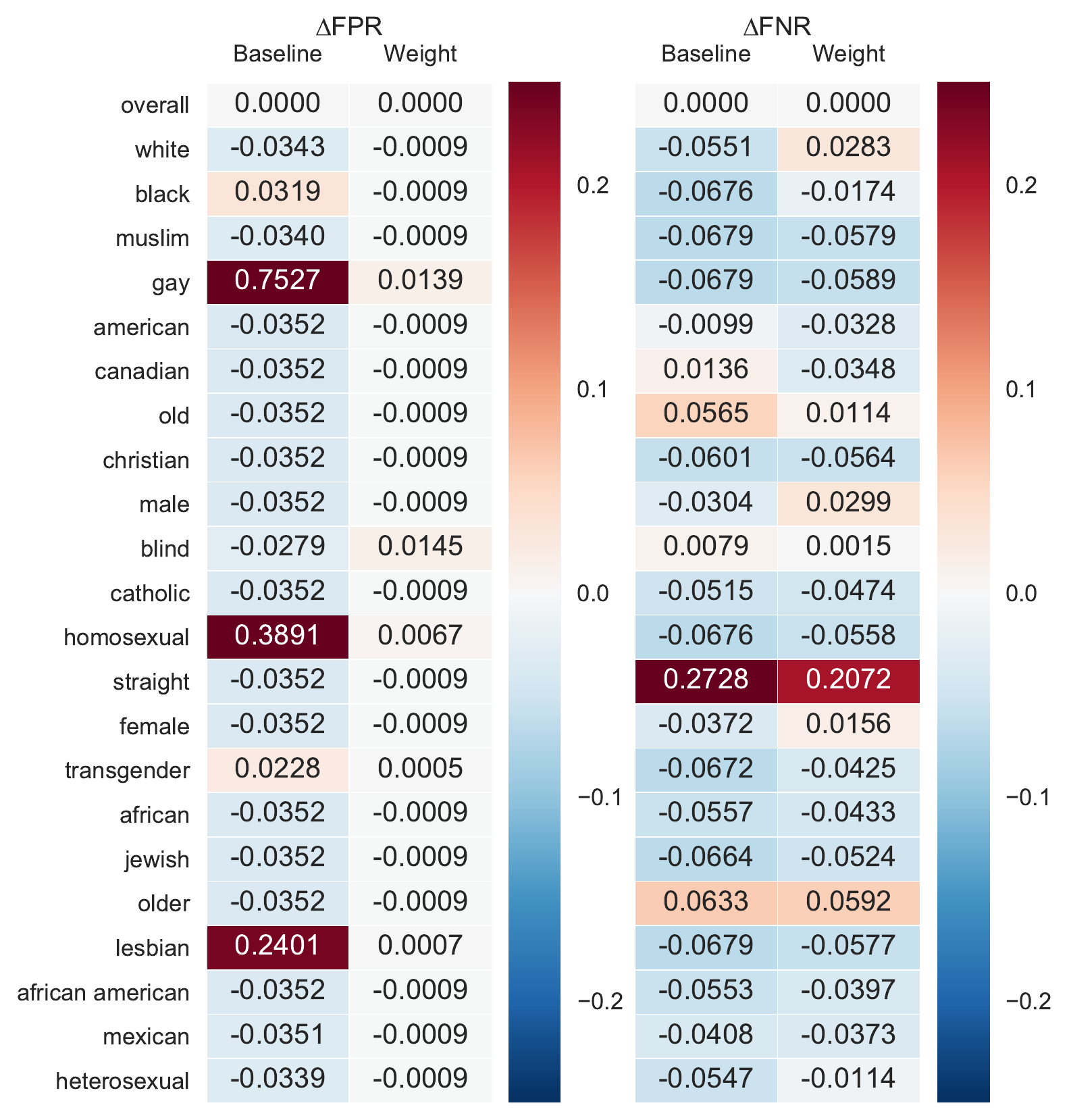}
  \caption{Comparison for the evaluation results of \emph{Baseline} and \emph{Weight} for sentences containing a selection of specific identities in IPTTS in Jigsaw Toxicity dataset, in which $\Delta \text{FPR}_{z}=\text{FPR}_{z}-\text{FPR}_\text{overall}$, and $\Delta \text{FNR}_{z}=\text{FNR}_{z}-\text{FNR}_\text{overall}$. Values closer to $0$ indicate better equality. Best viewed in color.}
  \label{fig}
\end{figure}

\paragraph{Jigsaw Toxicity}
Table~\ref{tab6} reports the results on Jigsaw Toxicity dataset. \textbf{\emph{Baseline}} refers to vanilla models. \textbf{\emph{Supplement}} refers to models trained and validated with $15249$ additional samples extracted from Toxicity Comments to balance the identity-terms across labels. \textbf{\emph{Weight}} refers to models trained with calculated weights.

Similar to results on Toxicity Comments, we find that both \emph{Weight} and \emph{Supplement} perform significantly better than \emph{Baseline} in terms of IPTTS AUC and FPED, and the results of \emph{Weight} and \emph{Supplement} are comparable.
On the other hand, we notice that \emph{Weight} and \emph{Supplement} improve FNED slightly, while the differences are not statistically significant at confidence level 0.05.

To gain better knowledge about the debiasing effects, we further visualize the evaluation results on the Jigsaw Toxic dataset for sentences containing some specific identity-terms in IPTTS in Figure~\ref{fig}, where $\Delta \text{FPR}_{z}$ and $\Delta \text{FNR}_{z}$ are presented. Based on the definition of FPED and FNED, values closer to $0$ indicate better equality.
We can find that \emph{Baseline} trained in the original biased dataset can discriminate against some demographic groups.
For example, sentences containing identity words like ``gay'', ``homosexual'' and ``lesbian'' are more likely to be falsely judged as ``toxic'' as indicated by $\Delta$FPR, while ones with words like ``straight'' are more likely to be falsely judged as ``not toxic'' as indicated by $\Delta$FNR.
We can also notice that \emph{Weight} performs more consistently among most identities in both FPR and FNR.
For instance, $\Delta$FPR of the debiased model in samples with ``gay'', ``homosexual'' and ``lesbian'' significantly come closer to $0$, while $|\Delta \text{FNR}|$ also drop for ``old'' and ``straight''.

We also note that $\text{FPR}_\text{overall}$ and $\text{FPR}_\text{overall}$ of \emph{Weight} are significantly better than the results of \emph{Baseline}, \emph{i.e.}, $\text{FPR}_\text{overall}$ results are $0.001$ and $0.035$ for \emph{Weight} and \emph{Baseline} respectively, and $\text{FNR}_\text{overall}$ results are $0.061$ and $0.068$ for \emph{Weight} and \emph{Baseline} respectively, representing that \emph{Weight} is both more accurate and more non-discriminatory on the IPTTS set.


\section{Conclusion}
\label{sec:conclusion}

In this paper, we focus on the unintended discrimination bias in existing text classification datasets.
We formalize the problem as a kind of selection bias from the \emph{non-discrimination distribution} to the \emph{discrimination distribution} and propose a debiasing training framework that does not require any extra resources or annotations.
Experiments show that our method can effectively alleviate discrimination.
It's worth mentioning that our method is general enough to be applied to other tasks, as the key idea is to obtain the loss on the \emph{non-discrimination distribution}, and we leave this to future works.

\section*{Acknowledgments}

Conghui Zhu and Tiejun Zhao are supported by National Key R\&D Program of China~(Project No. 2017YFB1002102).

\bibliography{acl2020}
\bibliographystyle{acl_natbib}

\appendix

\section{Proof for the Criterion Consistency Theorem}
\label{app:proof_thm_cc}

\begin{proof}

  Here we present the proof for Theorem~\ref{thm:eq}.

  For the Statistical Parity criterion,
  \begin{small}
    \begin{displaymath}
      \begin{split}
        \text{Pr}(\widehat{Y}|Z) &= \sum_{y \in [0,1]}{\text{Pr}(\widehat{Y},Y=y|Z)}\\
        &= \sum_{y \in [0,1]}{\text{Pr}(\widehat{Y}|Y=y,Z)\text{Pr}(Y=y|Z)}\\
        &= \sum_{y \in [0,1]}{\text{Pr}(\widehat{Y}|Y=y)\text{Pr}(Y=y)}\\
        &=\text{Pr}(\widehat{Y})\,\text{.}
      \end{split}
    \end{displaymath}
  \end{small}

  For the Predictive Parity criterion,
  \begin{small}
    \begin{displaymath}
      \begin{split}
        \text{Pr}(Y|\widehat{Y}=1,Z) &= \frac{\text{Pr}(\widehat{Y}=1|Y,Z)\text{Pr}(Y|Z)}{\text{Pr}(\widehat{Y}=1|Z)}\\
        &= \frac{\text{Pr}(\widehat{Y}=1|Y)\text{Pr}(Y)}{\text{Pr}(\widehat{Y}=1)}\\
        &= \text{Pr}(Y|\widehat{Y}=1)\,\text{.}
      \end{split}
    \end{displaymath}
  \end{small}
\end{proof}

\section{Discussion about Assumption~\ref{asmp:4}}
\label{app:pzqz}
We show that even if the assumption does not hold, we can still make models fit $Q(Y|X)$ with calculated weights when $Z$ is contained in $X$, which is the common setting in practical.

We firstly present the equation of the weights $w$ without the assumption $P(Z) = Q(Z)$.
\begin{small}
  \begin{equation*}
    \begin{split}
        w=&  \frac{Q(y)}{P(y|z)} \\
        \overset{1}{=}&  \frac{Q(y)}{Q(y|z, S=1)} \\
        \overset{bayes}{=}&  \frac{Q(y)}{Q(S=1|z,y) Q(y|z) / Q(S=1|z)}\\
        \overset{2,3}{=}&  \frac{Q(S=1|z)}{Q(S=1|x,z,y)}\\
        \overset{bayes}{=}&  \frac{Q(S=1)}{Q(S=1|x,z,y)}\cdot\frac{Q(z|S=1)}{Q(Z)}\\
        \overset{1, bayes}{=}&  \frac{Q(x,z,y)}{P(x,z,y)}\cdot\frac{P(z)}{Q(z)}\,\text{.}
    \end{split}
  \end{equation*}
\end{small}

After applying these weights to every sample in the dataset, we can get a new distribution defined as below,
\begin{small}
\begin{equation*}
 P^*(x,y,z_x) = \frac{w_{x,y,z_x} \cdot P(x,y,z_x)}{\int w_{x', y', z_x'} \cdot dP(x',y',z_x')}\,\text{.} \\
\end{equation*}
\end{small}
in which we use $P^*(\cdot)$ to represent the probability in the obtained distribution.
As $Z$ is contained in $X$, we use $Z_X$ to represent the specific $Z$ contained in every $X$.

Then we have
\begin{small}
    \begin{equation*}
      \begin{split}
        P^*(y|x) &= \frac{P^*(x,z_x,y)}{\sum_{y'}{P^*(x,z_x,y')}} \\
        &= \frac{P(x,z_x,y) \cdot \frac{Q(x,z_x,y)}{P(x,z_x,y)} \cdot \frac{P(z_x)}{Q(z_x)}}{\sum_{y'}{P(x,z_x,y') \cdot \frac{Q(x,z_x,y')}{P(x,z_x,y')} \cdot \frac{P(z_x)}{Q(z_x)}}}\\
        &= \frac{\frac{P(z_x)}{Q(z_x)} \cdot Q(x,z_x,y)}{\frac{P(z_x)}{Q(z_x)} \cdot \sum_{y'}{Q(x,z_x,y')}}\\
        &= Q(y|x,z_x)\\
        &= Q(y|x) \,\text{,}
      \end{split}
    \end{equation*}
    \end{small}
    and
    \begin{small}
        \begin{equation*}
      \begin{split}
        P^*(x) &= \sum_{y}{P^*(x,z_x,y)}\\
        &= \sum_{y}{\frac{w_{x,z_x,y} \cdot P(x,z_x,y)}{\int w_{x', z_{x'}, y'} \cdot dP(x',z_{x'},y')}}\\
        &\propto \sum_{y}{\frac{Q(x,z_x,y)}{P(x,z_x,y)} \cdot \frac{P(z_x)}{Q(z_x)} \cdot P(x,z_x,y)}\\
        &= \frac{P(z_x)}{Q(z_x)} \cdot Q(x, z_x)\\
        &= \frac{P(z_x)}{Q(z_x)} \cdot Q(x)\,\text{.}
      \end{split}
    \end{equation*}
\end{small}

Given the result $P^*(y|x) = Q(y|x)$, 
the consistent learners should be asymptotically immune to different assumptions regarding $Q(Z)$, 
where a learner is defined as consistent if the learning algorithm can find a model $\theta$ that is equivalent to the true model at producing class conditional probabilities given an exhaustive training data set~\cite{fan2005improved}.
In practical, however, as the requirements are often hard to met, we note that models may still be affected due to the deviation between $P^*(x)$ and $Q(x)$, which is widely studied in the covariate shift problem~\citep{shimodaira2000improving,ben2007analysis,jiang2007instance}.
In our paper, as we don't assume the availability of extra resources and prior knowledge, we simply set $P(Z)=Q(Z)$.
We leave more explorations about this assumption for future work.

\section{Details about the IPTTS Generation}
\label{app:iptts}

\begin{table}[!h]
  \centering
  \resizebox{0.482\textwidth}{!}{
    \begin{tabular}{c|c}
      \hline
      Templates                                     & Label \\
      \hline
      (verb inoffensive) (identity-term).           & 0     \\
      (verb offensive) (identity-term).             & 1     \\
      I am (identity-term).                         & 0     \\
      I hate (identity-term).                       & 1     \\
      Being (identity-term) is (adj. inoffensive).  & 0     \\
      Being (identity-term) is (adj. offensive).    & 1     \\
      You are a (adj. inoffensive) (identity-term). & 0     \\
      You are a (adj. offensive) (identity-term).   & 1     \\
      \hline
    \end{tabular}
  }
  \caption{Templates used to generate IPTTS.}
  \label{temp}
\end{table}

\begin{table}[!h]
  \centering
  \resizebox{0.482\textwidth}{!}{
    \begin{tabular}{c|c}
      \hline
      Type             & Example Words                     \\
      \hline
      adj. inoffensive & great, fun, nice, neat            \\
      adj. offensive   & disgusting, filthy, nasty, rotten \\
      verb inoffensive & hug, love, like, respect          \\
      verb offensive   & kill, murder, hate, destroy       \\
      male identity    & actor, airman, boy, man           \\
      female identity  & actress, airwoman, girl, woman    \\
      \hline
    \end{tabular}
  }
  \caption{Examples of slotted words to generate IPTTS.}
  \label{words}
\end{table}

For experiments with the Sexist Tweets dataset, we generate IPTTS following \citet{park2018reducing}.
The templates used are the same as \citet{park2018reducing}, as shown in Table~\ref{temp}.
We use the released codes by \citet{dixon2018measuring} and use the gender word pairs released by \citet{zhao2018gender}  as ``identity-term.''
Some of the slotted words are presented in Table~\ref{words}.
To make sentences longer, we also add some semantic-neutral sentences provided by \citet{dixon2018measuring} as a suffix to each template.
Finally, we get $76497$ samples, $38254$ of which are abusive, and the mean of sentence lengths is $17.5$.

For experiments with Toxicity Comments datasets and Jigsaw Toxicity, we use the IPTTS released by \citet{dixon2018measuring}.
The testing set is created by several templates slotted by a broad range of identity-terms, which consists of $77000$ examples, $50\%$ of which are toxic.

\section{Frequency of Identity-terms in Toxic Samples and Overall}

\begin{table}[!t]
  \centering
  \resizebox{0.482\textwidth}{!}{
    \begin{tabular}{c|c|c|c|c|c|c}
      \hline
      \multirow{2}*{Term} & \multicolumn{3}{c|}{Origin} & \multicolumn{3}{c}{Weight}                                                     \\
      \cline{2-7}
      ~                   & Toxic                       & Overall                    & $\Delta$       & Toxic & Overall & $\Delta$       \\
      \hline
      white               & 5.98                        & 2.13                       & 3.85           & 2.89  & 2.14    & \textbf{0.75}  \\
      black               & 3.10                        & 1.07                       & 2.03           & 1.22  & 1.07    & \textbf{0.15}  \\
      muslim              & 1.57                        & 0.58                       & 0.99           & 0.58  & 0.59    & \textbf{-0.01} \\
      gay                 & 1.29                        & 0.35                       & 0.94           & 0.39  & 0.34    & \textbf{0.05}  \\
      american            & 2.70                        & 2.11                       & \textbf{0.59}  & 2.76  & 2.13    & 0.63           \\
      canadian            & 1.38                        & 1.82                       & -0.44          & 1.48  & 1.82    & \textbf{-0.34} \\
      old                 & 2.62                        & 2.18                       & \textbf{0.44}  & 2.63  & 2.18    & 0.45           \\
      christian           & 0.89                        & 0.54                       & 0.35           & 0.73  & 0.55    & \textbf{0.18}  \\
      male                & 0.73                        & 0.44                       & 0.29           & 0.41  & 0.45    & \textbf{-0.04} \\
      blind               & 0.51                        & 0.28                       & \textbf{0.23}  & 0.55  & 0.28    & 0.27           \\
      catholic            & 0.63                        & 0.82                       & -0.19          & 0.65  & 0.83    & \textbf{-0.18} \\
      homosexual          & 0.26                        & 0.08                       & 0.18           & 0.09  & 0.08    & \textbf{0.01}  \\
      straight            & 0.51                        & 0.37                       & 0.14           & 0.46  & 0.38    & \textbf{0.08}  \\
      female              & 0.50                        & 0.37                       & 0.13           & 0.28  & 0.37    & \textbf{-0.09} \\
      transgender         & 0.21                        & 0.09                       & 0.12           & 0.10  & 0.09    & \textbf{0.01}  \\
      african             & 0.30                        & 0.20                       & 0.10           & 0.20  & 0.20    & \textbf{0.00}  \\
      jewish              & 0.28                        & 0.19                       & 0.09           & 0.17  & 0.19    & \textbf{-0.02} \\
      older               & 0.16                        & 0.25                       & \textbf{-0.09} & 0.15  & 0.25    & -0.10          \\
      lesbian             & 0.11                        & 0.03                       & 0.08           & 0.03  & 0.03    & \textbf{0.00}  \\
      african american    & 0.16                        & 0.09                       & 0.07           & 0.09  & 0.10    & \textbf{-0.01} \\
      mexican             & 0.20                        & 0.13                       & 0.07           & 0.17  & 0.13    & \textbf{0.04}  \\
      heterosexual        & 0.09                        & 0.03                       & 0.06           & 0.03  & 0.03    & \textbf{0.00}  \\
      \hline
    \end{tabular}}
  \caption{Frequency of a selection of identity-terms in toxic samples and overall in Jigsaw Toxicity dataset. \% is omitted.}
  \label{tab7}
\end{table}

To give a better understanding of how the weights change the distribution of datasets, we compare the original Jigsaw Toxicity dataset and the one with calculated weights for the frequency of a selection of identity-terms in toxic samples and overall, as shown in Table~\ref{tab7}.

We can find that after adding weights, the gap between frequency in toxic samples and overall significantly decrease for almost all identity-terms, which demonstrate that the unintended bias in datasets is effectively mitigated.

\end{document}